%% file: main.tex
\PassOptionsToPackage{usenames,dvipsnames}{xcolor}
\documentclass[sigconf]{acmart}

\usepackage{amsmath}
\usepackage{amssymb}
\usepackage{algorithm}
\usepackage{algorithmic}
\usepackage{multirow}
\usepackage{hhline}
\usepackage{url}            
\usepackage{booktabs}       
\usepackage{amsfonts}       
\usepackage{nicefrac}       
\usepackage{microtype}      
\usepackage{hhline}
\usepackage{ctable}
\usepackage{mathtools}
\usepackage{array}
\usepackage{multirow}
\usepackage[inline]{enumitem}
\usepackage{subfig}
\usepackage{multirow}
\usepackage{color}
\usepackage{siunitx}
\usepackage[font=small,labelfont=bf]{caption}

\DeclareGraphicsExtensions{.eps,.ps,.pdf,.png,.tiff,.jpg,.jpeg}

\DeclareMathAlphabet\mathbfcal{OMS}{cmsy}{b}{n}

\newcommand{\new}[1]{{\textcolor{black}  {#1}}}

\definecolor{redcol}{rgb}{1, 0, 0}
\definecolor{bluecol}{rgb}{0, 0, 1}
\newcommand{\red}[1]{\textcolor{redcol}{#1}} 

\renewcommand{\paragraph}[1]{\smallskip\noindent{\bf{#1}}}
\def\Appr{$\text{A}_\text{R}$}
\def\Semr{$\text{S}_\text{R}$}
\def\Appl{$\text{A}_\text{L}$}
\def\Seml{$\text{S}_\text{L}$}

\def\eg{\emph{e.g.}}
\def\ie{\emph{i.e.}}

\def \etal{\emph{et al.}}

\AtBeginDocument{%
  \providecommand\BibTeX{{%
    \normalfont B\kern-0.5em{\scshape i\kern-0.25em b}\kern-0.8em\TeX}}}

\copyrightyear{2020}
\acmYear{2020}
\setcopyright{acmlicensed}\acmConference[MM '20]{Proceedings of the 28th ACM International Conference on Multimedia}{October 12--16, 2020}{Seattle, WA, USA}
\acmBooktitle{Proceedings of the 28th ACM International Conference on Multimedia (MM '20), October 12--16, 2020, Seattle, WA, USA}
\acmPrice{15.00}
\acmDOI{10.1145/3394171.3413647}
\acmISBN{978-1-4503-7988-5/20/10}

\begin{document}
\fancyhead{}
\title[RGB2LIDAR: Towards Solving Large-Scale Cross-Modal Visual Localization]{RGB2LIDAR: Towards Solving Large-Scale Cross-Modal \\ Visual Localization}
%

\author{Niluthpol Chowdhury Mithun, Karan Sikka, Han-Pang Chiu, Supun Samarasekera, Rakesh Kumar}
\affiliation{%
  \institution{Center for Vision Technologies, SRI International, Princeton, NJ}
}
\email{ {niluthpol.mithun, karan.sikka, han-pang.chiu, supun.samarasekera, rakesh.kumar}@sri.com }

%

%
\begin{abstract}

We study an important, yet largely unexplored problem of large-scale cross-modal visual localization by matching ground RGB images to a geo-referenced aerial LIDAR 3D point cloud (rendered as depth images). Prior works were demonstrated on small datasets and did not lend themselves to scaling up for large-scale applications. To enable large-scale evaluation, we introduce a new dataset containing over $550K$ pairs (covering $143 km^2$ area)  of RGB and aerial LIDAR depth images. We propose a novel joint embedding based method that effectively combines the appearance and semantic cues from both modalities to handle drastic cross-modal variations. Experiments on the proposed dataset show that our model achieves a strong result of a median rank of $5$ in matching across a large test set of $50K$ location pairs collected from a $14km^2$ area. This represents a significant advancement over prior works in performance and scale. We conclude with qualitative results to highlight the challenging nature of this task and the benefits of the proposed model. Our work provides a foundation for further research in cross-modal visual localization.

\end{abstract}




\maketitle

\input{intro}

\input{related_works}

\input{approach}

\input{experiments}

\input{conclusions}
\begin{acks}
We would like to thank Ajay Divakaran for proof-reading the manuscript and providing many helpful comments. We would also like to acknowledge Zachary Seymour and Avi Ziskind for providing valuable suggestions and helping in preparing the dataset.
\end{acks}

\bibliographystyle{ACM-Reference-Format}
\bibliography{main}


\end{document}

%% file: intro.tex
\section{Introduction}

The real-world environment can be represented in many data modalities that are sensed by disparate sensing devices. For example, the same scene can be captured as an RGB or IR image, a depth map, or a set of 3D point clouds.  Due to significant growth in the availability and diversity of sensors, the problem of matching data across heterogeneous
modalities has gained noticeable momentum in recent years. Moreover, being able to
match an image to different data modalities, whose acquisition is simpler/faster, opens more opportunities for visual localization (\textbf{VL}) across different autonomous platforms, \eg, unmanned-ground vehicles. 


\begin{figure}[t!]
	\centering
	\vspace{0.3cm}
	\includegraphics[width=0.48\textwidth]{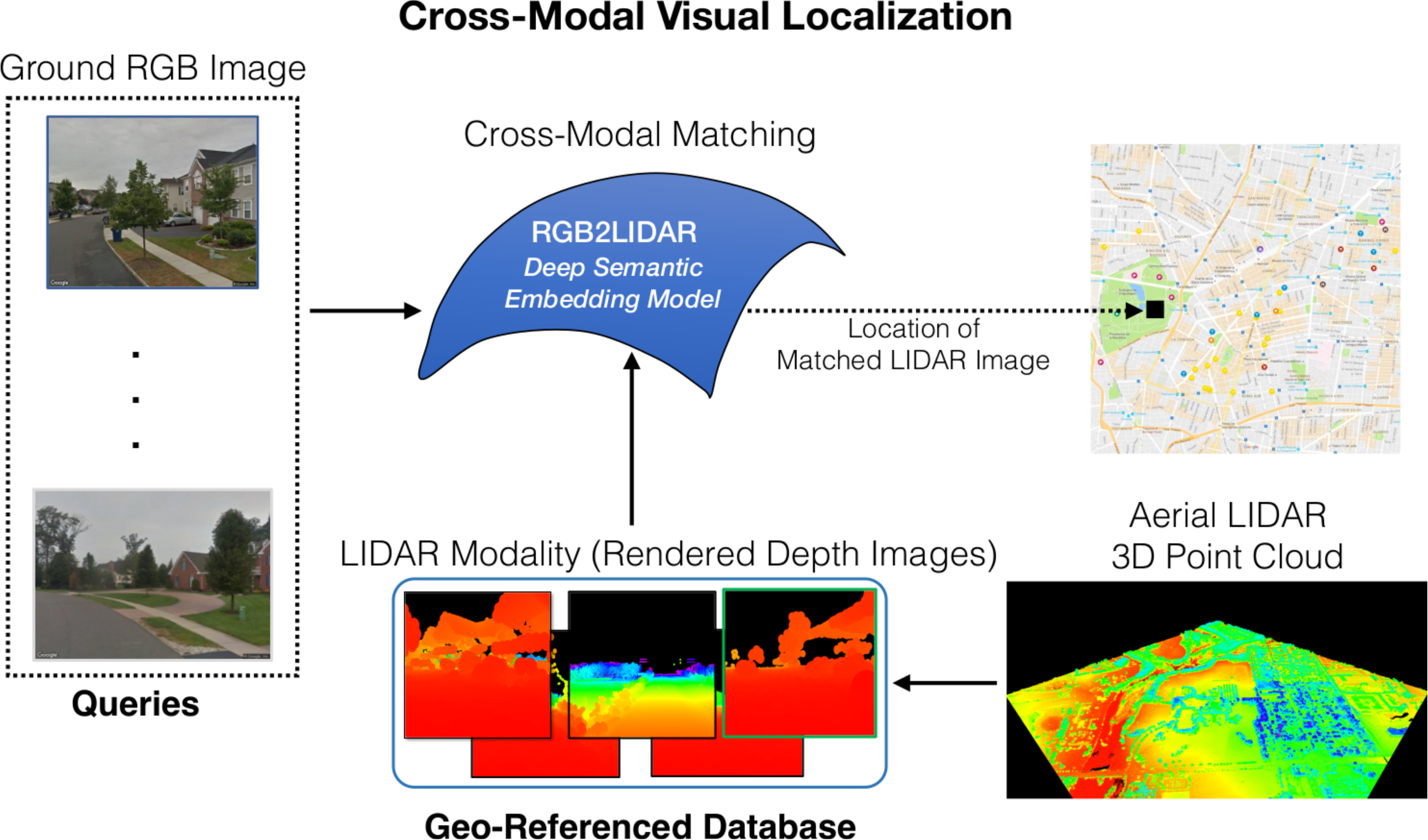}
	\caption{A brief illustration of our cross-modal geo-localization approach by matching a query ground RGB image to a geo-referenced aerial LIDAR 3D point cloud rendered as depth images.}
	\label{fig:problem}
	\vspace{-0.2cm}
\end{figure}

\vspace{0.05cm}
In this paper, we target the problem of matching instances across diverse data modalities in a large-scale setting.
Specifically, our goal is to localize a given street-level RGB image by querying a geo-referenced database from a
different data modality, \ie, LIDAR (Fig.~\ref{fig:problem}).  Prior works \cite{matei2013image, bansal2014geometric,
castaldo2015semantic} are limited in tackling this task since they (1) were evaluated on small benchmark
datasets and reported low accuracies, (2) heavily relied on hand-crafted features resulting in limited robustness to
cross-modal appearance variations, and (3) operated mostly in specific settings such as urban areas.
As a result, it has been difficult to establish the effectiveness of such works for real-world applications. 


\vspace{0.05cm}
We address this problem by contributing a learning-based approach \textbf{RGB2LIDAR} and creating a
large-scale dataset
, referred to as \textbf{GRAL}, to
evaluate large-scale cross-modal retrieval.
GRAL consists of $550K$ location-coupled cross-modal pairs (covering $143km^2$ area)
collected automatically from Google Street View and USGS LIDAR data based on location coordinates. 
We choose 3D point cloud data from a LIDAR sensor for constructing the geo-referenced database. LIDAR, that directly
senses the world in 3D, has recently become a popular data source for localization
\cite{huval2015empirical,wolcott2014visual,yu2015semantic}. We particularly examine data acquired from an airborne LIDAR
sensor.  This problem is challenging compared to matching images across time (ground-to-ground) \cite{chen2017deep, milford2015sequence,
toft2018semantic, mithun2018learning1, naseer2017semantics}, or viewpoint  changes in the same modality (ground-to-aerial) \cite{workman2015wide, lin2015learning, hu2018cvm, tian2017cross, regmi2019bridging}. Additionally, since LIDAR data points on
vertical surfaces are often missing from aerial collections \cite{matei2013image}, the matching problem becomes more
difficult. The benefit of using an aerial LIDAR collection system is that it covers a larger area much faster than the
traditional time-consuming collections from the ground. Thus, matching technologies between ground RGB images and aerial
LIDAR can benefit many VL applications where the geographical tags for images are unavailable, including historical
images, self-driving cars, and images taken from GPS-denied or GPS-challenged environments.  Instead of directly using
3D point clouds, we use LIDAR depth images generated from the projections of the point cloud data at a uniform grid of
2D locations on the ground plane. LIDAR data remains somewhat invariant to the projections in 2D and hence allows 
matching across modalities \cite{matei2013image}.  Moreover, it is easier to tackle large-scale visual localization in 2D as compared to
3D, since it is hard to work with 3D maps for large areas \cite{seymour2018semantically,sattler2017large}.

\vspace{0.05cm}
\new{The proposed approach (RGB2LIDAR VL) will enable autonomous vehicle-robot navigation. The traditional solution to this problem has been to use costly LIDAR sensors to localize the robot using geo-referenced LIDAR data in 3D world coordinates. The new trend is to use low-cost cameras (cf. LIDAR) to match with the geo-referenced 3D database, which requires 2.5D rendering  from the 3D database due to the difficulty of direct 2D-3D matching \cite{hirzer2017efficient, chiu2018augmented, arth2015instant}. In GPS-challenged environments, it typically involves two steps: (1) coarse search (or, geo-tagging \cite{matei2013image}) of the 2D input image to find a set of candidate matches of 2.5D maps from the database, (2) fine alignment performs 2D-3D match verification for each candidate and returns the estimated 3D (6-DoF) geo-pose. As long as the candidate list includes the correct match, the fine alignment step can estimate accurate 3D geo-pose. Our work is the first to provide a large-scale cross-modal RGB to aerial LIDAR based solution for the first step (i.e., coarse search). The potential benefit from this application (such as 3D alignment) cannot be enabled from image-to-image (either ground-to-ground or ground-to-aerial) localization.}

\vspace{0.05cm}
To the best of our knowledge, RGB2LIDAR is the first deep learning-based approach for large-scale cross-modal VL.  
The proposed method is based on utilizing multimodal deep convolutional neural networks (CNN) to learn joint representations 
for ground-level RGB images and aerial LIDAR depth images. This formulates cross-modal VL as a retrieval problem, that matches query RGB images to geo-referenced LIDAR depth images in the database
(Fig.~\ref{fig:problem}). Since the data from the two modalities exhibit large variations in appearances (Fig.~\ref{fig:dataset}), appearance information alone is not sufficient for identifying accurate correspondences. Semantic information can help significantly in this regard, as it focuses on the overall scene layout, and are generally more consistent as compared to appearance cues across
heterogeneous data modalities (Fig.~\ref{fig:segmentation_example})
\cite{arandjelovic2014visual,schonberger2018semantic, radwan2018vlocnet++, seymour2018semantically}. 
We use the representations generated from segmentation networks for LIDAR depth maps as semantic cues. However, it is difficult to obtain labels for training these models. 
We thus study and compare two complementary methods to tackle this issue. The first method is based on training a LIDAR segmentation network using cross-modal weak supervision from the segmentation maps of the paired RGB images from GRAL. The second involves training a segmentation network on a recently introduced DublinCity dataset that is the first semantic labeled dataset for dense aerial laser scanning \cite{zolanvari2019dublincity}.
To effectively fuse the appearance and semantic cues, we train multiple joint embedding models using different combinations of cues, and perform a weighted fusion. 

\vspace{0.05cm}
Our experiments show that the proposed RGB2LIDAR model performs significantly better than prior works and baselines
(Table~\ref{tab:pretrained}). Our appearance-only joint embedding model achieves a median rank of $19$ in matching
across a large test set, whereas the random baseline is $24,921$ (Table~\ref{tab:ablation}). \new{The proposed fusion with
semantic cues achieves median rank 5, a relative improvement of $73.6\%$ over the appearance-only model.}
We also highlight the benefits of using weak cross-modal supervision for obtaining semantic cues.  
We believe that
the underlying ideas and dataset developed in this work will open up avenues for further work in cross-modal VL.

\vspace{0.05cm}
\underline{\textit{Contributions: }} The main contributions of this work can be summarized as follows.

\vspace{-0.1cm}

\begin{enumerate}
	\item We study an important, yet largely unexplored problem of large-scale cross-modal visual localization.
	Prior works were demonstrated on small datasets and did not lend themselves to scaling up for large-scale applications. 
	We hope our work will encourage future work in cross-modal VL.
		
 \vspace{0.05cm}
		
	\item We propose, to the best of our knowledge, the first deep learning-based method for cross-modal VL. The proposed RGB2LIDAR method is based on training joint representations and simultaneously utilizing appearance and semantic cues from cross-modal pairs. 
	
	\vspace{0.05cm}
	
	\item To enable large-scale evaluation for the task of cross-modal VL, we introduce a new large-scale dataset
		containing $550K$ location-coupled cross-modal pairs of ground RGB images and rendered depth images from
		aerial LIDAR point cloud covering around $143 km^2$ area (cf.  $5 km ^2$ from \cite{matei2013image}).
 
 \vspace{0.05cm}
	
	\item We compare two complementary approaches for training the semantic segmentation network (used to obtain semantic cues) for LIDAR depth images-- first based on weakly supervised training of a LIDAR depth segmentation network. The second is based on training with full supervision from a dataset of limited	diversity.

\vspace{0.05cm}

	\item We perform extensive experimental studies to establish the challenging nature of this problem and show the advantages of the proposed model compared to prior works. The proposed model achieves a strong result of a median rank $5$  in matching across a large test set of $50K$ location pairs collected from a $14km^2$ area.
	

\end{enumerate}

%% file: related_works.tex
\vspace{0.05cm}
\section{Related Works}

\vspace{0.05cm}
\textbf{Visual Localization:} Vision-based methods generally localize
\cite{baatz2012leveraging, hays2008im2gps,li2012worldwide,sattler2011fast,
mithun2018learning1,schonberger2018semantic} or categorize
\cite{pronobis2006discriminative, rottmann2005semantic, wu2008place} a query
image based on a database of geo-referenced images or video streams
\cite{badino2012real, levinson2007map, arroyo2015towards}. Most prior works
consider the problem of visual localization by matching location-coupled image
database collected from the same viewpoint and sensor modality
(Electrical-Optical Camera) as the query image.  Significant work has been done
in recent years in this direction utilizing hand-designed feature descriptors
\cite{cummins2010fab, lategahn2013learn}, pre-trained CNN based features
\cite{chen2014convolutional,sunderhauf2015performance,sunderhauf2015place}, and
feature learning based approaches
\cite{naseer2017semantics,schonberger2018semantic,mithun2018learning1, zhu2018attention, seymour2018semantically}.
Although these methods show good localization performance, their applications 
are limited by the difficulty in collecting reference ground images
covering a large area.  We refer our readers to \cite{piasco2018survey},
\cite{lowry2016visual} and \cite{garcia2015vision} for a comprehensive
reviews on state-of-the-art approaches on vision-based localization.  

\vspace{0.05cm}
\textbf{Cross-View Visual Localization:}  To overcome the limitation of ground collections, several recent works have tried to localize ground-level image by matching against reference aerial imagery, which are easier to obtain \cite{lin2013cross, workman2015wide, lin2015learning, hu2018cvm, tian2017cross, regmi2019bridging,Shi2019spatial}. However, these cross-view localization approaches suffer from significant perspective distortion due to the drastic viewpoint changes. \new{The performance of these methods are also quite low when matching to single-view query images, and requires panoramic ground-view images as queries to achieve good accuracy.}

\vspace{0.05cm} 
\textbf{Cross-Modal Visual Localization:} Over the last
decade, we have seen significant growth in the availability and diversity of
sensors. Matching an image to different data modalities, that are simpler to be
collected, opens more opportunities and possibilities for large-scale visual
localization.  However, this problem has been rarely investigated and has
relied heavily on hand-crafted
features \cite{castaldo2015semantic,matei2013image,wang2015lost,piasco2018survey}.
In contrast to prior works, we propose the first deep learning method to tackle
this problem and demonstrate its feasibility for large-scale visual
localization.  

\vspace{0.05cm}
\textbf{Image-to-LIDAR Visual Localization:} Our data choice for cross-modal visual localization is matching ground RGB images to aerial geo-referenced LIDAR depth data \cite{matei2013image,bansal2014geometric}. In terms of the
experimental setup, our work is most closely related to \cite{matei2013image}. However, \cite{matei2013image} needs manually annotated building outlines for the query image during matching, which is not practical for large-scale applications. \cite{matei2013image}
also depends on local feature matching and urban scene to get distinctive features and is thus unlikely to work when paired images are not strongly aligned. A closely related work uses geometric point-ray features to compute candidate query poses without any appearance matching \cite{bansal2014geometric}. However, the method is also limited to urban settings and depends heavily on the availability of building corners in the image. Moreover, utilizing hand-crafted features limits the performance of these approaches \cite{bansal2014geometric, matei2013image}. Additionally, they are evaluated on very few queries and the reported accuracy is also quite low.



\vspace{0.05cm}
\textbf{Joint Embedding:} Joint embedding models have shown excellent performance on several multimedia tasks, \eg, cross-modal retrieval
\cite{wang2016learning, klein2015associating, huang2017learning, datta2019align2ground, wu2019unified, wu2019learning, wang2019matching, chen2019cross, mithun2018webly, mithun2019joint}, image
captioning \cite{mao2014deep,karpathy2015deep}, image classification
\cite{hubert2017learning, frome2013devise, gong2014multi} video summarization
\cite{choi2017textually,plummer2017enhancing}, cross-view matching
\cite{workman2015wide}. Cross-modal retrieval methods require computing
similarity between two different modalities, \eg, RGB and depth. Learning a
joint representation naturally fits our task of cross-modal RGB-LIDAR retrieval
since it is possible to directly compare RGB images and LIDAR depth images in
such a joint space.



%% file: approach.tex
\vspace{0.05cm}
\section{Approach}

\vspace{0.05cm}

In this section, we describe the proposed GRAL dataset and the RGB2LIDAR approach.

\vspace{-0.03cm}
\subsection{GRAL Dataset}
\vspace{-0.07cm}

\begin{figure}[t]
	\centering
	\vspace{0.2cm}
	\includegraphics[width=0.47\textwidth]{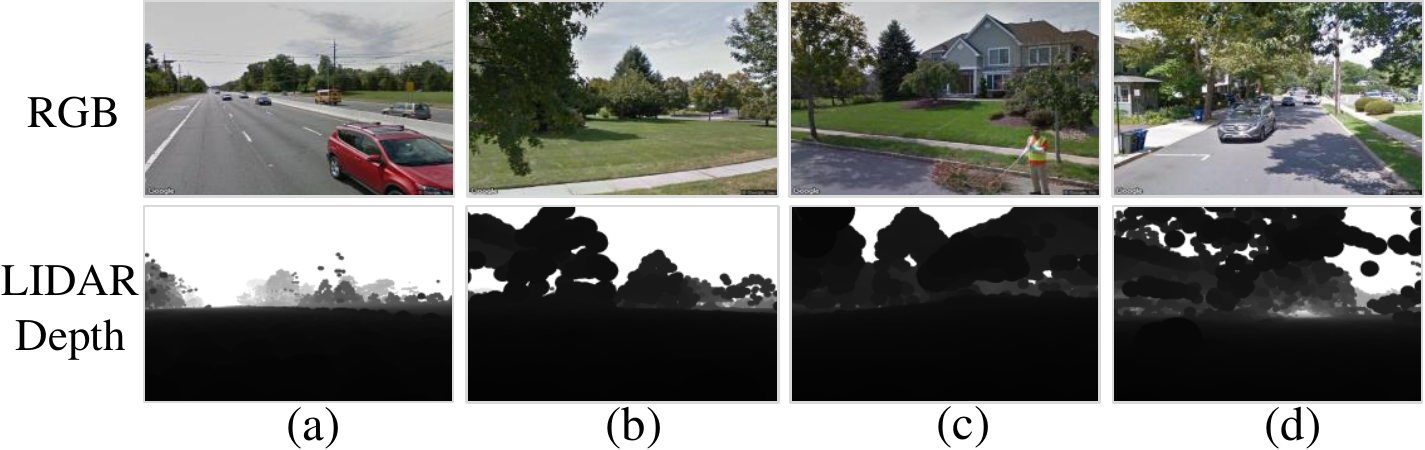}
	\vspace{-0.1cm}
	\caption{Example pairs of ground RGB images and aerial LIDAR depth images from the GRAL Dataset. RGB images are collected from Google Street View and depth images are collected by rendering aerial LIDAR point cloud from USGS.}
	\vspace{-0.2cm}
	\label{fig:dataset}
\end{figure}

A major obstacle in exploring the large-scale cross-modal visual localization task is that none of the existing datasets
are suitable for evaluation. To mitigate this issue, we create a new dataset which contains over $550K$
location-coupled pairs of ground RGB images and depth images collected from aerial LIDAR point clouds
(Fig.~\ref{fig:dataset}). We shall refer to this dataset as the Ground RGB to Aerial LIDAR (GRAL) dataset. Although the
primary purpose of this dataset is to evaluate cross-modal localization, it also allows us to evaluate matching under
challenging cross-view setting. See the project page \footnote{https://github.com/niluthpol/RGB2LIDAR} for additional details.

\subsubsection{\textbf{Data Collection}}
\vspace{-0.1cm}

We select a $143 km^2$ area around Princeton, NJ, USA for data collection. We choose the area as it contains diverse urban, suburban
and rural terrain characteristics. Our dataset contains a wide variety of scenes including forest, mountain, open country,
highway, city interior, building, street etc. Note that, the LIDAR to image geo-localization approach in \cite{matei2013image} also collected data from New Jersey but within a $5 km^2$ area.

\vspace{0.05cm}
To ensure that each ground RGB image is paired with a single distinct depth image from aerial LIDAR, we create our dataset in two phases. First, we download available ground RGB images in the selected area for different GPS
locations (latitude, longitude) using the Google Street View API. Second, we use LIDAR scan of the area from USGS to create a Digital Elevation Model (DEM) and from the DEM, location-coupled LIDAR depth images are collected for each
street view images. For each location, we used $12$ heading (\ang{0} to \ang{360} at \ang{30} intervals) for data collection.

\vspace{0.05cm}
\noindent \textbf{Harvesting Ground RGB Images:}
We collect ground RGB images by densely sampling GPS locations. As the images are only available on streets from Google
Street View, RGB imagery is not available in many locations in the selected area and Google returns a generic image for
these locations. We use image metadata from the street view API to filter these images. We ensure that selected
locations are $5$ meters apart as we empirically found that spacing the samples around $5$ meters apart tended to get a
new image. We ultimately list about $60K$ GPS coordinates on the streets for data collection. We hard set the image pixel size
($640\times480$), the horizontal field of view as $60$ and pitch as $0$ in the API. 

\vspace{0.05cm}

\noindent \textbf{Harvesting LIDAR Depth Images: }
We collect aerial LIDAR point-cloud of the selected area to create a DEM which is rendered exhaustively from multiple
locations and viewpoints. For each GPS location containing RGB images, we render the LIDAR depth images from $1.7m$
above the ground. A digital surface model is used for height correction above sea level. We remove the depth images with no
height correction and corresponding RGB images as we empirically found the viewpoint of these depth images are different from paired RGB images in most cases. We also remove the pairs where more than $60\%$ pixels are black in the depth image.

\vspace{0.05cm}
\new{Insufficient accuracy of instruments (GPS and IMU) and calibration issues may add some noise in data collection. 
Hence, there may exist some mis-alignment between location-coupled RGB and LIDAR depth images. To ensure good alignment between RGB and LIDAR depth images, we manually select a set of $100$ locations on the LIDAR point cloud and corresponding points from Google Earth and then calculated the best fitting offset to bring the LIDAR point cloud closest to the points on Google Earth. The offset is ($2.77m$ easting, $0.18m$ northing) in UTM coordinates. We consider this offset when rendering LIDAR depth images.}

\subsubsection{\textbf{Dataset Summary}}
Collecting readily available data from Google and USGS allowed us to create a large dataset without any manual effort.
\new{However, as the data collection process may introduce some mis-alignment in the dataset, we attempted to limit the issue by finding an offset based on a few manually selected points.} The dataset still may encounter small horizontal and vertical mis-alignment in some cases (\eg, (b) in Fig.~\ref{fig:dataset}). Moreover, as the LIDAR data collection is performed from an airborne sensor, the rendered depth images may miss many pixels (\eg, (b) in Fig.~\ref{fig:dataset}). A change in the scene is also observed in some cases due to the time difference in the data collection of Street View and LIDAR. The dataset also exhibits challenges common in most visual
localization datasets, such as dynamic aspects of environment (\eg, (a) in Fig.~\ref{fig:dataset}).

\vspace{0.05cm}
To evaluate how our approach generalizes to unseen areas, we hold out a part of images for testing and another part for validation. \new{Dataset split was done spatially. We use approximately $20\%$ of the area for collecting validation images, $10\%$ of the area for collecting testing images, and the rest for training.} The collected dataset finally contains $557,627$ location-coupled pairs with $417,998$ pairs for training, $89,787$ pairs for validation and $49,842$ pairs for the testing. For each unique location, we initially collected $12$ pairs based on different heading. However, after cleaning, the number of pairs for some locations may be less. 

\subsection{RGB2LIDAR}

We consider that during training we are provided with paired examples of RGB images and LIDAR depth images captured from
the same geo-location.  
During testing, we perform cross-modal retrieval to find the geo-location of a query RGB image
by matching it with a geo-referenced database of LIDAR depth images (Fig.~\ref{fig:problem}).

\vspace{0.05cm}
We first describe a general framework for cross-modal matching based on learning a joint multimodal embedding using
appearance information (Sec.~\ref{vse}).  This task is challenging since it involves matching examples across modalities
exhibiting large disparities in appearance characteristics. As a result, appearance information alone is not sufficient
for yielding high-quality matches.  In comparison to appearance information, higher-level scene information is generally
better preserved across inputs, from different visual sensors, capturing the same scene
\cite{arandjelovic2014visual,schonberger2018semantic, radwan2018vlocnet++, seymour2018semantically}. We thus propose to use semantic information
from intermediate feature maps generated from semantic segmentation networks for both modalities. 
We then discuss the proposed overall approach for fusing joint embeddings learned on different combination of appearance
and semantic cues for improved cross-modal matching (Sec.~\ref{matching}). A key issue with using semantic information
for LIDAR is that it is difficult to obtain or annotate semantic segmentation maps for LIDAR depth images. To tackle
this issue, we explore two approaches in Sec.~\ref{segmentation}. The first is based on using weak cross-modal supervision from the RGB modality to train a segmentation network for the
LIDAR depth images. The second approach is based on using representations from a segmentation
network for LIDAR depth images trained on a different (smaller) dataset with ground-truth labels. The two approaches are complementary in that
the first approach utilizes noisy annotations but has access to larger in-domain data. While the second approach uses a
cleaner data from a small possibly out-of-domain dataset.

\vspace{0.05cm}
\subsubsection{\textbf{Training Joint Multimodal Embedding}} \label{vse}
\vspace{-0.1cm}

We use a triplet ranking loss to embed both modalities in a common embedding space using the training pairs 
\cite{kiros2014unifying, zheng2017dual, wang2016learning, karpathy2014deep, FaghriFKF17}. We denote the feature vector of RGB image and LIDAR depth image as $\boldsymbol{f_r}$ ~($\in \mathbb{R}^I$) and
$\boldsymbol{f_d}$ ($\in \mathbb{R}^L$) respectively. We use a linear projection to project both modalities in a common
space-- $\boldsymbol{r}_p = W^{(r)}\boldsymbol{f_{r}}$~($\boldsymbol{r}_p \in \mathbb{R}^J$) and $\boldsymbol{d}_p =
W^{(d)}\boldsymbol{f_d}~(\boldsymbol{d}_p \in \mathbb{R}^J)$. 
Here, $W^{(r)} \in \mathbb{R}^{J \times I}$ and $W^{(d)}\in \mathbb{R}^{J \times L}$ project RGB and LIDAR depth maps to the joint space respectively. Using pairs of feature representation of RGB images and corresponding depth images, the goal is to learn a joint
embedding such that the pairs from similar geo-locations, \ie, positive pairs are closer that than negative pairs in the feature space. We achieve this by using a bi-directional triplet ranking loss as shown below:
\begin{equation} \label{eq:equation1}
\begin{split}
	\mathcal{L}_p =& \max_{\hat{\boldsymbol{d}}_p}[\Delta -S(\boldsymbol{r}_p, \boldsymbol{d}_p) + S(\boldsymbol{r}_p, \hat{\boldsymbol{d}}_p)]_{+} \\  
	+& \max_{\hat{\boldsymbol{r}}_p}[\Delta -S(\boldsymbol{r}_p, \boldsymbol{d}_p) + S(\hat{\boldsymbol{r}}_p, \boldsymbol{d}_p)]_{+} 
\end{split}
\end{equation}
where $[x]_{+} = \max(x, 0)$, $\mathcal{L}_p$ is the loss for a positive pair $(\boldsymbol{r}_p, \boldsymbol{d}_p)$, 
$\hat{\boldsymbol{r}}_p$ and $\hat{\boldsymbol{d}}_p$ are the negative samples for $\boldsymbol{r}_p$ and
$\boldsymbol{d}_p$ respectively. $\Delta$ is the margin value for the loss. 
We use cosine similarity as the scoring function $S(\boldsymbol{r}_p,\boldsymbol{d}_p)$.
We sample the negatives in a stochastic manner from each minibatch. 

\vspace{0.05cm}
We utilize this framework to learn multimodal embeddings for different combinations of appearance and semantic features
from the two modalities. 
We rely on using features from the semantic segmentation networks trained for both RGB and LIDAR depth images
for semantic information.  
Although there are prior works on segmenting point clouds \cite{wang2018pointseg, hackel2017semantic3d, landrieu2018large}, they are not
directly applicable to aerial LIDAR depth images in our setting. 
We later discuss two approaches for obtaining segmentations from LIDAR depth maps. 

\subsubsection{\textbf{Combining Appearance and Semantic Cues}} \label{matching}

In order to exploit both the appearance and semantic information effectively, we construct our retrieval system based on a model ensemble \cite{polikar2007bootstrap,fraz2012ensemble}, where multiple expert models are generated strategically and
combined to obtain a high-quality predictor. As the success of ensemble approaches depends significantly on the
diversity of models inside the ensemble \cite{polikar2007bootstrap}, it is important for us to choose a diverse set of joint embeddings. We propose to use both appearance and semantic features by training four joint embedding models utilizing different combinations of these features from the RGB and the LIDAR depth images. We train these models using the multimodal embedding framework discussed in Sec.~\ref{vse}. 

At the time of retrieval, given a query ground image $\boldsymbol{r}$, the similarity score of the query is computed in each of the joint embedding spaces with LIDAR images $\boldsymbol{d}$ from the dataset. Our mixture of expert
model uses a weighted fusion of scores for the final ranking.  

\vspace{-0.1cm}
\begin{equation}
\begin{aligned}
\label{eq:equation6}
\hspace{0.2cm} S(\boldsymbol{r},\boldsymbol{d})=\boldsymbol{w_{1}}S_{App-App}(\boldsymbol{r},\boldsymbol{d}) + \boldsymbol{w_{2}}S_{App-Sem}(\boldsymbol{r},\boldsymbol{d})~+ \\ \boldsymbol{w_{3}}S_{Sem-App}(\boldsymbol{r},\boldsymbol{d}) + \boldsymbol{w_{4}}S_{Sem-Sem}(\boldsymbol{r},\boldsymbol{d})
\end{aligned}
\end{equation}
where hyphen (-) symbol in subscript below $S$ separates features from RGB and LIDAR depth used in learning joint the representations. For example, $S_{App-Sem}$ refers to the similarity score calculated in the joint space trained with appearance feature from RGB and semantic feature from LIDAR depth.
The values of  $\boldsymbol{w_{1}}$, $\boldsymbol{w_{2}}$, $\boldsymbol{w_{3}}$, $\boldsymbol{w_{4}}$ are chosen empirically based on the validation set.

\vspace{0.05cm}
In this work, we adopt a simple fusion strategy since our focus is to investigate the importance of utilizing multiple cues together for the task and whether it leads to a significant improvement over using only one of the cues. We later show in experiments that such a strategy helps in improving results highlighting the complementarity of the embedding models. We believe more sophisticated fusion strategies (\eg, \cite{mahmood2019structured, zadeh2017tensor}) will further improve the performance of our retrieval model. We leave this as future work. 


\subsubsection{\textbf{Semantic Cues for LIDAR Depth Images}}
\label{segmentation}


We use feature maps generated from pre-trained segmentation networks from each input modality as the semantic cues. For
RGB modality, we use a pre-trained segmentation network \cite{badrinarayanan2015segnet}, which works 
reasonably well. \new{However, to the best of our knowledge, there exists no pre-trained segmentation network for aerial LIDAR point-clouds.} Moreover, it is extremely difficult to obtain manual annotations for training segmentation networks on LIDAR depth images. We also observe that directly utilizing features from networks trained on RGB images for extracting semantic cues from LIDAR images and then training joint embeddings results in poor performance. We explore two complementary
approaches to tackle this issue.

\vspace{0.05cm}
\noindent \textbf{Weak Cross-Modal Supervision: } The first method is inspired by the success of supervision transfer from paired multi-modal data in several vision and multimedia tasks
\cite{gupta2016cross, xu2017learning, christoudias2010learning}. 
Our method is motivated by two simple intuitions- (1) although there exist no pre-trained segmentation networks for aerial
LIDAR depth images, there exist powerful segmentation networks for RGB images, and (2) the RGB and the LIDAR depth
images are weakly aligned, and thus the RGB segmentation maps contain useful information about the general layout of the
captured scene. We utilize the segmentation maps extracted from the paired RGB image as the ground-truth maps to train
the segmentation network for LIDAR depth maps. We believe that due to the weak alignment between the modalities, the RGB
segmentation maps contain sufficiently rich signals to train a reasonable segmentation network for the LIDAR depth
images. Further, deep networks have been shown to be robust to noisy ground-truth and having some ability to
self-correct their estimate of the noisy ground-truth in tasks such as weakly supervised segmentation and 
unsupervised machine translations \cite{lample2018phrase, khoreva2017simple}. 

\vspace{0.05cm}
\noindent \textbf{Full supervision from a smaller dataset: } A possible drawback with the first approach is the use of noisy
annotations from RGB modality, which are only weakly aligned with the LIDAR modality. This could result in a weak
model that may not generalize well. We also explore a second approach where we use features from a segmentation network
trained on a different dataset. We use the DublinCity dataset which contains labeled LIDAR point clouds collected around
Dublin city centre and it is the first labeled dataset for a dense Aerial Laser Scanning \cite{zolanvari2019dublincity}.
Although the dataset provides clean labels for training, the coverage area is small ($2km^2$) and thus allows to collect
a limited diversity of scenes.  Hence, despite using cleaner labels the segmentation network is only trained with a
smaller diversity of images and could contain out-of-domain examples, which might not transfer well. 

\vspace{0.05cm}
We later show these complementary approaches are able to provide useful semantic cues for retrieval,
highlighting the need for using alternate forms of supervision for cross-modal localization. 

%% file: experiments.tex
\section{Experiments}
\label{sec:exp}

We now evaluate our method on the task of cross-modal VL by retrieving aerial LIDAR depth images by using the ground RGB
images as queries. We first describe the evaluation metric and the implementation details. Then, we establish useful
baselines on the proposed dataset and highlight the challenging nature of the task and the dataset.  We position our
work relative to the prior works on cross-modal VL through a discussion due to the difficulty of an empirical comparison
since these methods used nonstandard datasets and pipelines. Next, we perform an analysis of our model to establish the
benefits of the proposed fusion of appearance and semantic cues for cross-modal retrieval. We also evaluate different
approaches for training the segmentation networks for LIDAR depth images and provide useful insights for future
works. We finally present some qualitative results.

\vspace{-0.05cm}
\paragraph{Evaluation Metric:} We use standard evaluation metrics used in prior cross-modal retrieval tasks~\cite{kiros2014unifying,dong2016word2visualvec, mithun2018learning}. We report R@K (Recall at K) that calculates the percentage of queries for which the ground truth (GT) results are found within the top-$K$ retrievals (higher is better). 5m R@1 calculates the percentage of queries for which the best matching sample is found within $5$-meter distance to the GT. We also report MedR which calculates the median rank of the GT results in the retrieval (lower is better).

\vspace{-0.05cm}
\paragraph{Implementation Details:}  The joint embedding models are trained using a two-branched neural network consisting of expert networks for encoding each modality and corresponding fully-connected layers to transform their outputs to a joint (aligned) representation ($J=1024$) \cite{wang2016learning,kiros2014unifying}.
We initialize appearance and semantic CNN models for both modalities using networks pre-trained for image classification and
semantic segmentation respectively. We use an $18$ layer wide-ResNet model \cite{zagoruyko2016wide} pre-trained on Places365 \cite{zhou2017places} as the RGB appearance CNN.
We initialize the depth appearance model
with pre-trained weights from the same appearance CNN, which we find works reasonably well in experiments. We use PSPNet model trained on ADE20K \cite{zhou2019semantic} as the semantic segmentation network for RGB \cite{zhao2017pyramid}. We also attempted initializing depth semantic CNN with pre-trained RGB semantic CNN models. However, as the retrieval performance was quite low, we train LIDAR depth semantic segmentation models based on the approaches described in Sec.~\ref{segmentation}. We use SegNet model \cite{badrinarayanan2015segnet} trained on Cityscapes \cite{Cordts2016Cityscapes} to initialize the LIDAR depth segmentation network. 


%
%



\subsection{Baselines on GRAL dataset} 

Previous methods for this problem typically relied on hand-crafted features \cite{matei2013image, bansal2014geometric}.
On the other hand, pre-trained CNN features has been shown to be a strong baseline in many vision and multimedia tasks
\cite{Razavian2014cnn}. In Table~\ref{tab:pretrained}, we compare RGB2LIDAR with hand-crafted GIST, pre-trained
wide-ResNet18, and ResNet50 feature-based matching. Although directly utilizing hand-crafted or deep features for
matching is unlikely to be effective in our setting due to viewpoint and modality differences, we provide these
comparisons to understand the difficulty of the cross-modal (and cross-view) localization task on the proposed dataset. From
Table~\ref{tab:pretrained}, we observe that GIST (R@1 of $0.002\%$), wide-ResNet18 ($0.014\%$) and ResNet50 ($0.007\%$)
perform slightly better than chance ($0.002\%$), whereas the proposed RGB2LIDAR approach shows strong results ($27.6\%$).



\vspace{0.05cm}
We also consider predicting depth cues from RGB images, and then performing retrieval based on the predicted depth.
Specifically, we predict depth cues from RGB images using a single-view depth prediction model-- MegaDepth
\cite{li2018megadepth} and train joint spaces with appearance cues from LIDAR depth images. This achieves a significantly lower
performance compared to our RGB2LIDAR model ($0.9\%$ vs. $27.6\%$ of ours in R@1). 

\renewcommand{\arraystretch}{1.05}
\begin{table}[t]
	\caption{This table compares our RGB2LIDAR model with hand-crafted and pre-trained CNN features based matching for cross-modal localization on the GRAL dataset. The results highlight the difficulty of the task, where hand-crafted features and deep features only perform slightly better than chance. }
	\vspace{-0.2cm}
	\centering
	\small
	\scalebox{1}{
	\begin{tabular}{|@{\hspace{2.3\tabcolsep}} l @{\hspace{2.3\tabcolsep}}|@{\hspace{2.3\tabcolsep}} c @{\hspace{2.3\tabcolsep}} c @{\hspace{2.3\tabcolsep}} c @{\hspace{2.3\tabcolsep}} c @{\hspace{2.3\tabcolsep}} c @{\hspace{2.3\tabcolsep}}|}
		\hline
		Method        & R@1  & R@5  & R@10 & MedR & MeanR \\
		\hhline{|======|}
		Chance        & $0.002$ & $0.010$ & $0.020$ & $24921$ & $24921$  \\
		\hline
		GIST          & $0.002$ & $0.016$ & $0.026$ & $21101$ & $22479$  \\
		\hline
		wide-ResNet18 & $0.014$ & $0.059$ & $0.114$ & $19028$ & $20131$  \\
		\hline
		ResNet50      & $0.007$ & $0.031$ & $0.067$ & $19887$ & $20328$  \\
		\hline
		MegaDepth      & $0.9$ & $3.2$ & $5.1$ & $1735$ & $5628$  \\
		\hline
		\textbf{RGB2LIDAR}          & $\mathbf{27.6}$ & $\mathbf{51.1}$ & $\mathbf{57.9}$ & $\mathbf{5}$   & $\mathbf{34.5}$ \\
		\hline
\end{tabular}}
	\vspace{-0.3cm}
	\label{tab:pretrained}
\end{table}


\subsection{Comparison with Prior Work}


\paragraph{Ground RGB to Aerial Lidar based Retrieval: }
It is difficult to directly compare our approach with prior works on cross-modal localization \cite{bansal2014geometric,
	matei2013image} since the query images and the exact location of LIDAR data are not available.  Moreover, these methods
use specific pre-processing which makes them unfit for large-scale applications as targeted in our work. For example,
they are limited to an urban setting and localization performance depends on the availability of the buildings in the image. Moreover, \cite{matei2013image} also requires manually annotating the building outlines of query images.

\vspace{0.05cm}
Bansal \etal \cite{bansal2014geometric} evaluated their approach on $50$ Google Street View image queries and reported
only $20\%$ accuracy in $5m$ localization in the top-1000 ranks in $1Km \times 0.5Km$ area. On the other hand, our method
shows $34\%$ accuracy in $5m$ localization in top-1 rank based on testing across $50K$ pairs. Matei \etal
\cite{matei2013image} evaluated their approach on $14$ queries in $5km^2$ area and reported R@1 of $7\%$, whereas
our method shows R@1 of $27.6\%$ based on $50K$ queries in $14km^2$ area. Hence, our method is likely to be more effective
and generalizable than these methods \cite{bansal2014geometric,matei2013image}.


\paragraph{Ground-Aerial RGB based Retrieval: } 
\new{One of the motivations behind this work is that the performance of present ground-aerial methods is quite low in practice due to significant perspective distortions from viewpoint changes. Moreover, state-of-the-art ground-aerial methods require panorama image as queries for decent performance \cite{hu2018cvm}, whereas we use images with a $60$ degree field of view in our evaluation. We perform comparison with a prominent cross-view localization model CVM-Net-I \cite{hu2018cvm}, by collecting ground panoramas and aerial satellite images for locations in GRAL. We follow ground-aerial image dataset CVUSA \cite{workman2015wide} data collection protocol, which was used to train CVM-Net-I. We find that CVM-Net-I model achieves low accuracy (i.e., R@1 =$0.7\%$, R@5 =$2.9\%$, R@10 = $5.1\%$) in RGB$\rightarrow$Aerial-image based localization, whereas our RGB2LIDAR model achieves significantly better performance (i.e., R@1 =$27.6\%$, R@5 =$51.1\%$, R@10 = $57.9\%$) in RGB$\rightarrow$LIDAR as shown in Table~\ref{tab:pretrained}. Hence, we believe the proposed method is promising as a large-scale visual localization solution.}

\renewcommand{\arraystretch}{1.05}
\begin{table}[t]
	\caption{\small{Performance of the proposed RGB2LIDAR fusion of appearance and semantic cues for the cross-modal VL task on the GRAL dataset. We divide the table into two blocks to compare different RGB2LIDAR methods, i.e.,  joint embedding with different RGB and LIDAR features (2.1) and fusion of embeddings (2.2). \Appr~and~\Semr~refers to appearance and semantic features from RGB images respectively. The (+) symbol denotes the ensemble of embeddings. The (-) symbol separates features from RGB images and LIDAR depth images, \eg, ~\Appr-\Seml~method refers to the embedding learned using appearance feature from RGB images and semantic feature from LIDAR depth images.}}
	\vspace{-0.3cm}
	\small
	\begin{center}
		\begin{tabular}{|@{\hspace{0.6\tabcolsep}} m{0.45cm} @{\hspace{0.5\tabcolsep}} |@{\hspace{0.5\tabcolsep}} c @{\hspace{0.5\tabcolsep}} |@{\hspace{0.6\tabcolsep}} c @{\hspace{0.6\tabcolsep}} c @{\hspace{0.5\tabcolsep}} c @{\hspace{0.5\tabcolsep}} c @{\hspace{0.6\tabcolsep}} c @{\hspace{0.6\tabcolsep}}|}
			\hline
			& \multirow{2}{*}{Method}                 &  \multicolumn{5}{c|}{\underline{Evaluation Metric}}              \\
			& & R@1  & R@5   & R@10 & MedR  & 5m R@1  \\ \hhline{|=======|}
			\multirow{5}{*}{2.1} & Chance & $0.002$ & $0.01$ & $0.02$ & $24921$ & $0.003$ \\ & \Appr-\Appl                                 &  $20.3  $  & $39.0 $   & $45.1 $  & $19$  & $26.4$                  \\
			& \Semr-\Seml & $10.6 $  & $26.8 $ & $34.8 $  & $38$  & $14.1$                  \\
			& \Appr-\Seml                                & $9.5 $   & $24.3 $ & $32.0 $    & $48$  & $12.6$                  \\
			& \Semr-\Appl            & $18.6 $  &  $37.2 $ & $43.6 $  & $22$  & $24.4$                  \\  \hline
			\multirow{4}{*}{2.2} & \Appr-\Appl + \Semr-\Appl                & $24.8 $  & $45.5 $ & $51.8 $  & 9   & $31.5$                  \\
			& \Appr-\Appl+\Appr-\Seml          & $22.9 $  & $44.7 $ & $52.0 $    & $9$   & $29.2$                  \\
			& \Appr-\Appl+\Semr-\Appl+\Appr-\Seml          & $27.0  $  & $49.5 $ & $56.2 $  & $6$   & $34.0$                  \\
			& \bf{\Appr-\Appl+\Semr-\Appl+\Appr-\Seml+\Semr-\Seml}   & $\mathbf{27.6 }$  & $\mathbf{51.1 }$ & $\mathbf{57.9 }$  & $\mathbf{5}$   & $\mathbf{34.5}$     \\
			& (\textbf{Proposed}) & & & & & \\ 
			\hline            
		\end{tabular}
	\end{center}
	\label{tab:ablation}
	\vspace{-0.2cm}
\end{table}

\vspace{-0.1cm}
\subsection{Analysis of the Proposed Method-- RGB2LIDAR}

We analyze the performance of our approach in Table~\ref{tab:ablation} to show the benefits of the proposed ideas-
(i) joint embeddings, (ii) semantic information, and (iii) overall approach using weighted ensemble-based fusion of multiple
embeddings (RGB2LIDAR), for cross-modal retrieval. We use semantic features from the best performing segmentation model
trained with weak cross-modal supervision (Table~\ref{tab:semantic}). We divide the table into two blocks (2.1-2.2) to aid our study.

\begin{figure*}[t]
	\centering
	\includegraphics[width=0.98\textwidth]{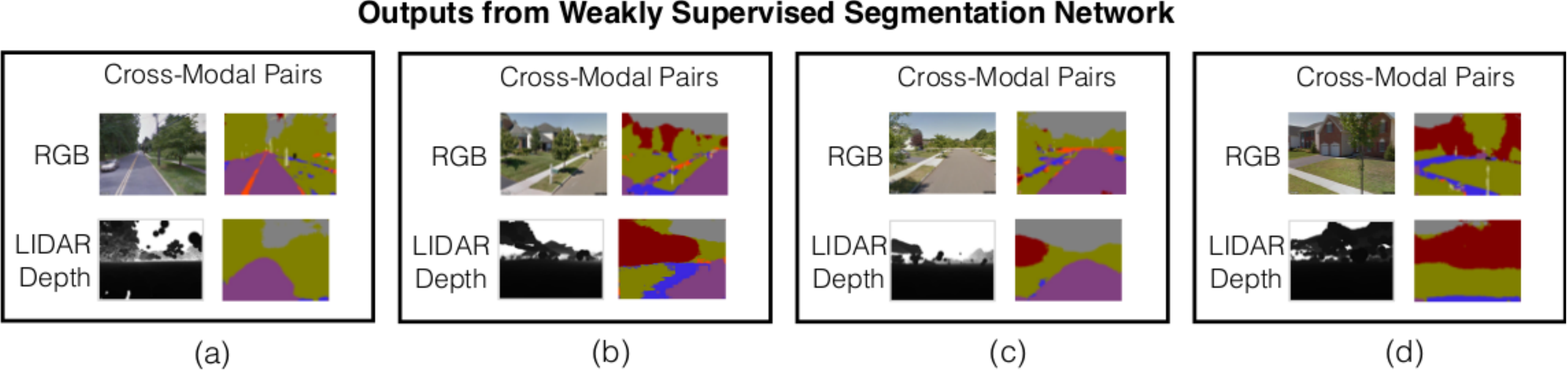}
	\vspace{-0.2cm}
	\caption{Sample semantic segmentation output for four cross-modal pairs. Here, pre-trained SegNet model is used for RGB image segmentation and segmentation network trained with weak cross-modal supervision is used for LIDAR depth image segmentation. The palette for semantic segmentation is as follows: \colorbox{lightgray}{sky,} \colorbox{BrickRed}{building,} \colorbox{GreenYellow}{pole,} \colorbox{orange}{road marking,} \colorbox{Plum}{road,} \colorbox{blue}{pavement,} \colorbox{LimeGreen}{tree}. Best viewed in color.}
	\vspace{-0.2cm}
	\label{fig:segmentation_example}
\end{figure*}

\paragraph{Different Image Features for Joint Embedding:} We first analyze the different combinations of appearance and semantic features from RGB and LIDAR depth images in training the joint embeddings in block-2.1. We observe that all four combinations perform reasonably well compared to the chance baseline. Utilizing the appearance features from both modalities seems to be performing the best. We note a drop in performance when using the semantic features, instead of the appearance features, in learning the joint embedding (R@1 is $20.3\%$ for \Appr-\Appl, $9.5\%$ for \Appr-\Seml and $18.6\%$ for \Semr-\Appl). We expect a higher drop for the case when using semantic features for the LIDAR depth images since the semantic segmentation network for the LIDAR modality is trained with noisy labels. Moreover, the rendered depth images are generally misaligned relative to their RGB counterparts leading to further noise in the generated
segmentations (Fig.~\ref{fig:dataset}). On the contrary, the drop in performance while using semantic features for
the RGB modality is not significant ($-1.7\%$ absolute in R@1 for \Semr-\Appl).

\paragraph{Fusion of Appearance and Semantics:} 
We propose a weighted fusion-based strategy to effectively combine the multiple joint embeddings (Sec.~\ref{matching}). 
\new{The fusion weights (Eq.~\ref{eq:equation6}) were $\boldsymbol{w_{1}}=0.33$, $\boldsymbol{w_{2}}=0.16$, $\boldsymbol{w_{3}}=0.32$, $\boldsymbol{w_{4}}=0.19$ and chosen based on a validation set.} 
The results from our weighted fusion-based approach, as shown in block-2.2, show large improvements over single joint embedding models (block-2.1). The proposed retrieval model utilizing both appearance and semantic cues achieves a $7.3\%$ absolute improvement in R@1 compared to the best performing single cue embedding model \Appr-\Appl.  We also observe improvements in fusion with the two joint embedding models trained using semantic cues from the LIDAR depth
images (\ie, \Appr-\Seml, \Semr-\Seml). For example, \Appr-\Appl + \Appr-\Seml shows an absolute R@1 improvement of $2.6\%$ over \Appr-\Appl.  We also observe improvements when comparing \Appr-\Appl + \Semr-\Appl~to the proposed model ($+2.8\%$ absolute in R@1). We also verify that the improvements do not occur by chance by averaging performance across $12$ randomized runs (R@1 of $27.37 \pm 0.24$ for proposed vs. $24.53 \pm 0.19$ when not using LIDAR semantic cues).

\vspace{0.05cm}
\new{We also performed experiments with two typical early fusion techniques (i.e., concatenation, max). We find the performance of these baseline approaches are significantly lower (R@1 is $20.4$ and $21.7$ respectively) than our approach. We believe this happens since these early fusion strategies are unable to preserve the intra-modal invariances captured by appearance and semantic cues in the final (fused) representation. }

\renewcommand{\arraystretch}{1.2}
\begin{table}[t]
\vspace{0.1cm}
	\centering
	\caption{Cross-modal retrieval performance with different semantic feature extracted from different
		segmentation networks for LIDAR depth images in learning joint embeddings. The results show that the semantic segmentation network trained using weak cross-modal supervision performs better.}
	\vspace{-0.2cm}
	\small
	\begin{center}
		\begin{tabular}{|@{\hspace{0.7\tabcolsep}} m{2.95cm} @{\hspace{0.8\tabcolsep}}|c @{\hspace{0.8\tabcolsep}}| c @{\hspace{0.9\tabcolsep}} c @{\hspace{0.9\tabcolsep}} c @{\hspace{0.9\tabcolsep}} c @{\hspace{0.9\tabcolsep}}|} \hline 
			\multirow{2}{*}{\parbox{2.9cm}{\centering LIDAR Semantic Feature (Supervision - Dataset)}} & \multirow{2}{*}{\parbox{1.4cm}{\centering RGB Image Feature}} & \multicolumn{4}{c|}{\underline{Evaluation Metric}}  \\
			&                 & R@1  & R@5  & R@10 & MedR     \\ \hhline{|=|=|====|}
			\multirow{2}{*}{\parbox{2.9cm}{\centering  SegNet-RGB\\ (Full - Cityscapes) }}    & Appearance                          & $1.9$  & $3.9$  & $5.8$  & $1958$       \\
			&          Semantic                         & $1.7$  & $4.1$  & $6.3$  & $1716$        \\ \hline
			\multirow{2}{*}{\parbox{2.9cm}{\centering SegNet-Depth \\ (Full - DublinCity)}}    & Appearance                          & $8.8$  & $22.0$  & $30.7$  & $52$       \\
			&      Semantic                         & $9.0$  & $23.4$  & $30.9$  & $53$        \\ \hline
			\multirow{2}{*}{\parbox{2.9cm}{\centering \textbf{wSegNet-Depth \\ (Weak - GRAL)}}}    & Appearance                             & $\mathbf{9.5}$  & $\mathbf{24.3}$ & $\mathbf{32.0}$   & $\mathbf{48}$       \\
			&        Semantic                         & $\mathbf{10.6}$ & $\mathbf{26.8}$ & $\mathbf{34.8}$ & $\mathbf{38}$    \\  \hline
		\end{tabular}
	\end{center}
	\label{tab:semantic}
	\vspace{-0.2cm}
\end{table}

\vspace{-0.05cm}
\subsection{Analysis of the LIDAR Depth Image Semantic Segmentation Network}
\label{sec:exp_sem}

We use feature maps from the outputs of LIDAR depth segmentation network for semantic features.  Since there are no ground-truth annotations, we first followed the standard practice of initialization with a pre-trained segmentation network on RGB images \cite{badrinarayanan2015segnet, wang2018depth}.  We observe in Table~\ref{tab:semantic} that the retrieval performance of this model is quite low ($1.9\%$ in R@1) highlighting that the representations are not directly transferable owing to the large domain (modality) gap. We now empirically discuss the two complementary approaches introduced in Sec~\ref{segmentation}. The first approach (wSegNet-Depth) is trained using weak supervision
from paired RGB images, while the second (SegNet-Depth) is trained using a labeled LIDAR depth point cloud dataset covering a smaller area.



\begin{figure*}[t]
	\centering
	\includegraphics[width=0.98\textwidth]{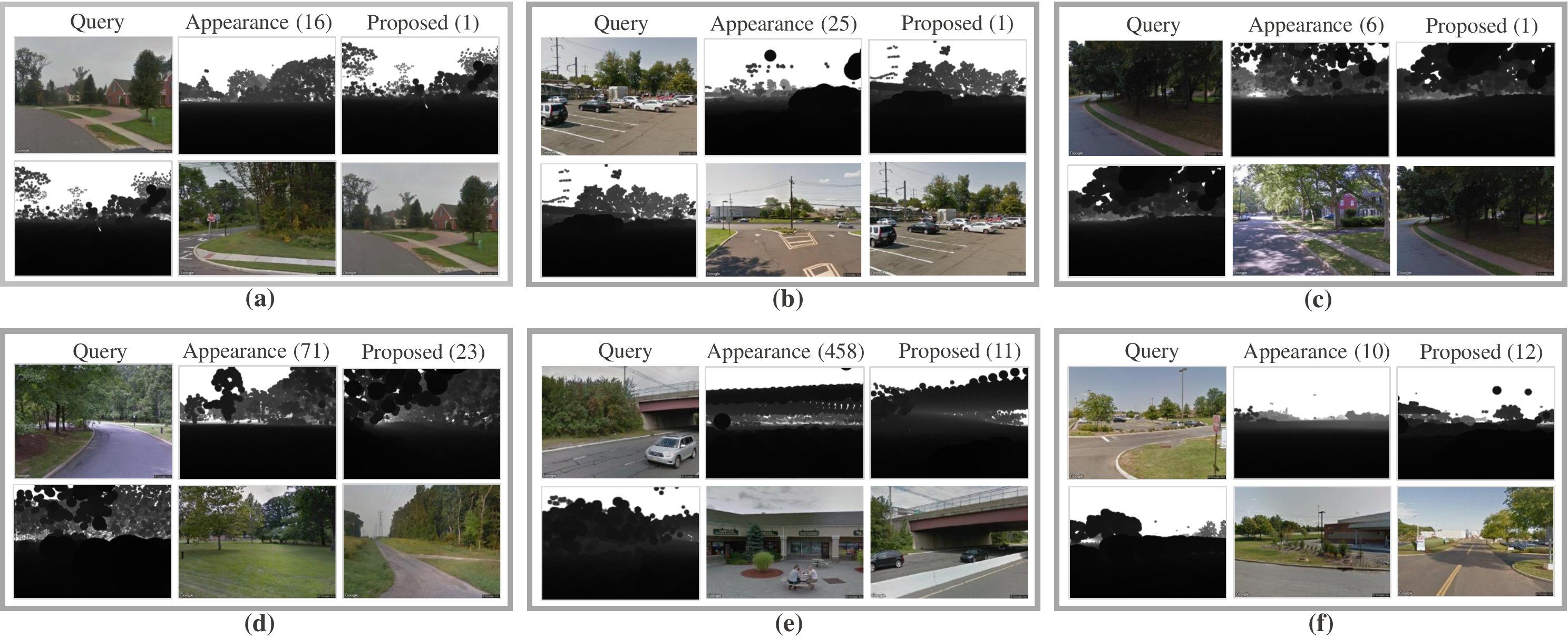} 
	\vspace{-0.1cm}
	\caption{Showing six query RGB images and top-1 retrieved LIDAR depth image obtained using appearance-only embeddings and the proposed method (fusing appearance and semantic cues). We also show the corresponding cross-modal pairs below each image and the rank of the ground-truth depth image in brackets for each method. \new{We observe both the proposed RGB2LIDAR method works well in retrieving the ground-truth match from a large set of about $50K$ candidates with good accuracy.}}
	\vspace{-0.2cm}
	\label{fig:qualit}
\end{figure*}

\vspace{0.05cm}
Table~\ref{tab:semantic} shows that both the SegNet-Depth and wSegNet-Depth model significantly outperform the
segmentation network initialized from pre-trained RGB segmentation network.  We also observe that the SegNet-Depth
model, despite being trained with noisy supervision, performs slightly better than the SegNet-Depth trained on
DublinCity (\eg $10.6\%$ vs. $9.0\%$ R@1 with semantic features from RGB).  We believe this happens since the
SegNet-Depth is trained on out-of-domain images collected in a smaller area and is thus not able to generalize as well
as the wSegNet-Depth, which is trained on in-domain examples covering a larger area.  However, the performance
gap is not huge given that SegNet-Depth is trained on a smaller area ($2km^2$ vs $101km^2$ in GRAL), which shows the
advantages of using clean labeled data. Prior works on learning from noisy supervision have also shown that large
quantities of noisy labels are required to match the performance of clean labels \cite{sun2017revisiting,
ahn2018learning, chen2015webly, yalniz2019billion}. We obtain further improvements by fine-tuning the weakly supervised
model with clean labeled data (\eg, R@1 of $28.2\%$ compared $27.6\%$ reported in Table~\ref{tab:ablation}).  
We believe that although the performance of cross-modal VL will improve as more labeled LIDAR point clouds
become available, there is potential in combining weak and full supervision for this task.  We also hope that our
findings will motivate future works in the exploration of freely available weak cross-modal supervision for handling multiple
modalities for this task. We show segmentation outputs for four LIDAR depth images generated using the wSegNet-Depth
segmentation network in Fig.~\ref{fig:segmentation_example}.

\subsection{Qualitative Results}

In Fig.~\ref{fig:qualit}, we show six cases of query ground RGB images along with the top-1 retrieved LIDAR depth images using RGB2LIDAR and the appearance-only embedding baseline. These results highlight the challenging nature of this task
especially the high visual similarity between different urban scenes, which necessitates the need for representations
that are discriminative yet invariant to cross-modal appearance variations \cite{schonberger2018semantic}. Overall, we observe both the proposed model and the appearance-only embedding baseline works well in retrieving the ground-truth match from a large set of about $50K$ candidates with good accuracy. However, our proposed fusion is able to consistently perform better in most cases. We expect this since cross-modal matching needs to deal with major changes in appearance across modalities, and semantic properties of a scene are generally more invariant to such factors. For example, in (e) the appearance-only model performs poorly in
retrieving the ground truth as the appearance cues seem to confuse the bridge with the building outline, while the
combined model is able to resolve this ambiguity. Since the segmentation of LIDAR depth images is not perfect, the
proposed model could sometimes degrade retrieval performance, \eg, (f).

%% file: conclusions.tex
\section{Conclusion}

\vspace{0.05cm}

We propose a new approach for large-scale cross-modal visual localization by matching query ground RGB images to a
geo-referenced depth image database constructed from aerial LIDAR 3D point cloud. The proposed RGB2LIDAR approach is
based upon a deep embedding based framework that capitalizes only on automatically collected cross-modal
location-coupled pairs in training and simultaneously utilizes appearance and semantic information by using an ensemble approach for efficient retrieval. We introduce a new dataset GRAL to evaluate the large-scale cross-modal VL task. 
We also compare two complementary approaches, that do not require any labels from our dataset, for training
the segmentation network (used to obtain semantic cues) for LIDAR depth images. 
The proposed approach is expected to generalize
well to unseen locations based on the strong results of a median rank $5$ in matching across a test set of about $50K$
location candidates covering an area of $14 km^2$. The underlying ideas developed in this work can be applied to matching RGB
images to other data choices for localization, \eg, 3D point clouds from other sensors, CAD model. In the future,
cross-modal visual localization could be a useful primitive across autonomous platforms. The proposed approach takes a
crucial step towards realizing that vision.